\pdfoutput=1

\documentclass[11pt]{article}

\usepackage[]{acl}

\usepackage{times}
\usepackage{latexsym}
\usepackage{dirtytalk}
\usepackage{adjustbox}
\usepackage{graphicx}
\graphicspath{ {./images/} }
\usepackage{hyperref}

\newcommand\blfootnote[1]{%
  \begingroup
  \renewcommand\thefootnote{}\footnote{#1}%
  \addtocounter{footnote}{-1}%
  \endgroup
}

\usepackage{xcolor}     
\usepackage{lipsum}     
\usepackage{ntheorem}   
\usepackage{mdframed}   

\theoremstyle{break}
\theoremheaderfont{\bfseries}
\newmdtheoremenv[%
linecolor=gray,leftmargin=60,%
rightmargin=40,
backgroundcolor=gray!40,%
innertopmargin=0pt,%
ntheorem]{myprop}{Proposition}[section]

\usepackage[T1]{fontenc}

\usepackage[utf8]{inputenc}

\usepackage{microtype}

\title{User Simulation with Large Language Models for\\Evaluating Task-Oriented Dialogue}

\author{Sam Davidson$^{1*}$, Salvatore Romeo$^{2\Diamond}$,  Raphael Shu$^{2\Diamond}$, James Gung$^2$,\\ \textbf{Arshit Gupta$^2$, Saab Mansour$^2$, Yi Zhang$^2$} \\$^1$ Department of Linguistics, University of California, Davis \\$^2$AWS AI Labs \\
\texttt{ssdavidson@ucdavis.edu}
\\ \texttt{\{romeosr, zhongzhu, gungj, arshig, saabm, yizhngn\}@amazon.com}}

\begin{document}
\maketitle
\begin{abstract}
\blfootnote{$^{*}$Work conducted during an internship at Amazon.}
\blfootnote{$^{\Diamond}$Corresponding author.}
One of the major impediments to the development of new task-oriented dialogue (TOD) systems is the need for human evaluation at multiple stages and iterations of the development process. In an effort to move toward automated evaluation of TOD, we propose a novel user simulator built using recently developed large pretrained language models (LLMs). In order to increase the linguistic diversity of our system relative to the related previous work, we do not fine-tune the LLMs used by our system on existing TOD datasets; rather we use in-context learning to prompt the LLMs to generate robust and linguistically diverse output with the goal of simulating the behavior of human interlocutors. Unlike previous work, which sought to maximize goal success rate (GSR) as the primary metric of simulator performance, our goal is a system which achieves a GSR similar to that observed in human interactions with TOD systems. Using this approach, our current simulator is effectively able to interact with several TOD systems, especially on single-intent conversational goals, while generating lexically and syntactically diverse output relative to previous simulators that rely upon fine-tuned models. 
Finally, we collect a Human2Bot dataset of humans interacting with the same TOD systems with which we experimented in order to better quantify these achievements.

\end{abstract}

\section{Introduction}

Automated evaluation of newly developed Task-Oriented Dialogue (TOD) systems remains a largely unsolved task in the NLP community. Two primary methods of evaluation have historically been used to test the quality of new TOD systems: dataset-based evaluation and interactive evaluation. Dataset-based evaluation, which is far more common in the literature, consists of using a TOD system to generate responses to annotated conversation logs from a fixed dataset, such as MultiWOZ \citep{Budzianowski2018MultiWOZA} or the Schema-Guided Dialogue (SGD) dataset \citep{Rastogi2020TowardsSM}. While long used as a benchmark in the TOD community, this type of evaluation fails to capture one key aspect of TOD system usage: the interactive exchange between a TOD system and its users. That is, when testing using benchmark data, the user utterances are fixed and thus cannot adapt to reasonable, though unanticipated, system responses. As discussed in \citet{Cheng2022IsMA}, the key weakness of this type of evaluation is the possibility of a policy mismatch between the generated TOD system utterance and the benchmark user utterance. In such a scenario, even reasonable and coherent responses by the TOD system may be penalized as they do not match the output expected in the benchmark dataset. Additionally, available TOD benchmark datasets are created using a predefined dialogue schema that limits the domains and intents on which a new TOD system can be tested using benchmark data. For example, the SGD Dataset \citet{Rastogi2020TowardsSM} is limited to twenty common domains, such as banking and weather. This shortcoming reduces the utility of dataset-based evaluation when testing TOD systems developed for novel domains.

Interactive evaluation, on the other hand, consists of a user simulator interacting with the target TOD system in a turn-by-turn manner, similar to a real conversation with a human user. Interactive evaluation benefits from the ability of the user simulator to adaptively respond to TOD-generated utterances, thereby reducing the likelihood of a policy mismatch between the system-side and simulated user-side utterances. This evaluation setup, while potentially more representative of the type of interaction a TOD system is likely to see in deployment, has heretofore been limited by the difficulty of creating an effective user simulator. However, recent work, such as that of \citet{Sun2022MetaphoricalUS} and \citet{Cheng2022IsMA}, has substantially advanced the art in TOD user simulator development. But this recent work still depends on large language models (LLMs) that have been fine-tuned on specific TOD benchmark datasets, thereby limiting the generalizability of the current generation of user simulators to unseen domains. 

In this paper, we present a user simulator which generates user-side utterances in a TOD conversation using only in-context learning with LLMs, thereby avoiding the fine-tuning stage of system development and removing the requirement of domain-specific conversation data for fine-tuning. In the context of developing a TOD system for a novel domain, the availability of such conversation data cannot be taken for granted. By removing the fine-tuning stage from our proposed user simulator's development pipeline, we are able to avoid any dependencies on extant domain-specific data. To engage in a conversations, our proposed user simulator requires only a set of predefined goals aligned to the schema of the target TOD system; these goals can be presented in either logical or natural-language form, and thus can be hand-written or automatically generated from the defined dialogue schema. While no fine-tuning of the underlying LLM is required in our system setup, simulator performance will depend on choice of LLM; we primarily experiment using GPTNeoX-20B \citep{black-etal-2022-gpt} as the basis for our user simulator, but have also tested our proposed in-context learning setup using GPT3 \citep{NEURIPS2020_1457c0d6}, BLOOM \citep{scao2022bloom}, and AlexaTM-20B \citep{soltan2022alexatm}.

We ask three primary research questions in this paper:

\begin{itemize}
\item Can in-context learning be effectively used to leverage the power of LLMs for simulating a user interacting with a TOD system?

\item How do we evaluate such a user simulator to demonstrate that it is generating diverse, “human-like” utterances, and that it is adaptable across domains?

\item How do these metrics, as calculated on a user simulator, compare to those metrics calculated on data of humans interacting with deployed TOD systems?

\end{itemize}

As stated above, our goal is to develop a user simulator that generates linguistically diverse, complex utterances that approximate real-world user input to TOD systems. We use automated and human evaluation to demonstrate the user simulator's ability to generate appropriate and diverse output in many conversational settings, which may be used to test in-development and deployed TOD systems. We collect a dataset of humans interacting with two state-of-the-art TOD systems, which we use to establish a baseline for human-level diversity and goal success. Finally, we present and discuss the shortcomings and limitations of the proposed in-context learning-based user simulator, most importantly the decline in system performance with more complex conversational goals, and discuss possible future directions to improve upon and mitigate these limitations.

\section{Related Work}

 Much of the previous work in user and dialogue simulation for TOD systems has focused on data augmentation for system training. For example, \citet{Mohapatra2021SimulatedCF} uses dialogues fully generated by fine-tuned LLMs for data augmentation in low-resource training scenarios. Similarly, \citet{chen2021teaching} generates complete dialogues in an effort to bootstrap training on unseen domains. This focus on using user simulators to generate training data follows a trend in past user-simulator work, much of which focused on creating new data for reinforcement learning (for example \citet{shi-etal-2019-build}. Thus, despite work such as \citet{LopezCzar2007TestingTP} and \citet{schatzmann2006survey}, the use of user simulators to test extant dialogue systems is a major gap in the literature, that may be due to the fact that existing models have until recently struggled to generate output that is both reasonable and linguistically diverse \citep{chen2021teaching}.

More recent work, such as \citet{Sun2022MetaphoricalUS}, \citet{Cheng2022IsMA}, and \citet{lin2022gentus}, has begun to close the this gap in the development of end-to-end user simulators for the evaluation of TOD systems. \citet{Sun2022MetaphoricalUS} presents an effective user simulator based on the T5 language model \citep{raffel2020exploring}, as well as a framework for testing TOD user simulators. Similarly, \citet{Cheng2022IsMA} build a user simulator by fine-tuning T5 on data from MultiWOZ. Likewise, \citet{lin2022gentus} used a fine-tuned version of BART \citep{lewis2020bart} as the basis of its TOD user simulator, which is further refined using reinforcement learning. While these simulators achieve impressive goal-success rates (GSR), their high GSR may actually be a negative in the context of user simulation, as discussed further below. In addition, none of the recent works harness the power of the more recent generation of LLMs (such as GPT3, ChatGPT, OPT, and GPTNeoX), nor do they take advantage of recent work in in-context learning (for example, \citet{kojima2022large}; \citet{wei2022chain}). Rather these models rely on fine-tuning pretrained language models on annotated, domain-specific dialogue data, thereby limiting their applicability in the testing of new TOD systems for previously unseen domains. While \citet{lin2022gentus} does report a reasonable degree of adaptability to unseen domains, the need for annotated data to fine-tune the system remains a potential issue when dealing with new dialogue schemata that diverge significantly from the schema used in creating the training data.

As discussed in \citet{li2021evaluate}, user simulators, if done right, can be a powerful means of evaluating new and in-development dialogue systems. In the present paper, our goal is to harness the power of the newer generation of LLMs along with recent developments in in-context learning to generate linguistically diverse, human-like user-side utterances that can be used to test and evaluate extant dialogue systems in a realistic setting. While we achieve this goal to a limited extent, we also find specific difficulties in the use of in-context learning for TOD user simulation, especially with more complex grounding goals.

\section{Evaluation Metrics}

We use several metrics to quantify the interaction of both human users and user simulators with extant TOD systems. In all cases, our goal in developing our user simulator is not maximizing these metrics, but rather building a system that simulates human-level performance.

\subsection{Goal Success Rate}
\label{sec:gsr}
In recent work using user simulators to evaluate TOD systems, such as \citet{Cheng2022IsMA}, \citet{Sun2022MetaphoricalUS}, and \citet{chen2021teaching}, the primary metric used to gauge the performance of user simulators is Goal Success Rate (GSR) also know as Task Success Rate (TSR). In this metric, every dialogue is scored with a 1 or 0 depending on whether or not the final dialogue state satisfies all of the slot requests in the original goal. While this is a reasonable metric for evaluating the performance of TOD systems, as we want TOD systems to be able to complete all of a user's requested tasks, it is an imperfect metric for measuring the performance of user simulators, especially when maximizing GSR becomes a measure of system success. Specifically, a user simulator that is trained to maximize GSR may result in a simulator which communicates with a TOD system in an artificial manner designed to maximize the efficiency of information exchange. When designing a system to simulate real-world human interaction with a TOD system, our goal is not to maximize the ability of the simulator to convey information to the TOD system with which it is interacting. Rather, we wish to build a simulator that communicates in a naturalistic manner. As such, our goal is to achieve a GSR comparable to that of human interlocutors speaking to the TOD system, as this \say{GSR correlation} indicates a simulator that provides information to the TOD system with efficiency similar to humans. After all, our end goal in developing a user simulator is creating a system that is able to test the abilities of a deployed or in-development TOD system in a manner similar to how the system will be challenged by human users \say{in the wild.} For this reason, in this work we do not seek to maximize GSR; rather we aim to create a system that achieves a reasonable GSR that is in line with what we see in our collected human data.

\subsection{Diversity Metrics}
\label{sec:diversity_metrics}

We propose using a set of metrics to quantify the lexical and syntactic diversity of our proposed user simulator. \citet{Sun2022MetaphoricalUS} has proposed using Distinct score \cite{Li2016ADO} to measure lexical diversity in user simulator output; however, this measure is essentially an n-gram version of type-token ratio (TTR), and thus is likely to suffer from the well-known issues with TTR, namely that TTR is not normalized for text length and is known to be unreliable for comparing dialogues of varying length \citep{richards1987type, ttr_issues}. To avoid these issues relative to lexical diversity, and to take into account issues of syntactic complexity and diversity, we adopt a set of metrics frequently used in the language acquisition and linguistic analysis research, namely:

        \begin{itemize}

                \item MTLD: Measure of Textual Lexical Diversity, or MTLD, \citep{mccarthy2010mtld}, is a modified version of TTR which is normalized for length effects. As mentioned above, TTR is well known to be impacted by length of the target text; that is, as texts increase in length the TTR begins to level out or even drop \citep{richards1987type}. MTLD addresses this issue and has been shown effective in measuring the lexical diversity of texts as short as 100 tokens \citep{koizumi2012effects}. Thus, MTLD allows us to consistently measure lexical diversity at both dialogue and dataset levels.
                
                \item Mean dependency distance: Mean dependency distance, the average number of words spanned by dependency links in a given text, is a well-established measure of syntactic diversity \citep{oya2011syntactic}, used for measuring cognitive load \citep{roark2007syntactic} as well as language development \citep{ouyang2022dependency}.
                
                \item  Dependency distance standard deviation: We report the standard deviation in the dependency distance of our user utterances as a proposed metric of syntactic diversity, as used in \citet{liang2017dependency}.
            
        \end{itemize}
        
\section{Human2Bot Dataset}

As previously mentioned, GSR alone is not a reasonable metric for evaluating the performance of user simulators. Rather, we argue that GSR must be considered relative to the goal success seen in human interaction with deployed TOD systems. Additionally, we wished to have a baseline against which to compare the lexical and syntactic diversity of output from our user simulator as well as those presented in previous work. To that end, we engaged eight human subjects to converse with two deployed TOD systems: PPTOD \citep{su2021multi} and Soloist \citep{peng2020soloist}. As with our user simulator, we provided the participants with goal information drawn from MultiWOZ, and asked them to use these goals to engage in conversations with the target TOD system. In addition to the goal information, subjects were asked to include a series of conversational complexities, such as providing two values for a single slot or changing a previously stated slot value at a later point in the conversation. These complexities encourage diversity in our collected dataset and serve to provide reference data for similar human behavior \say{in the wild}.

All human participants are native English speakers who regularly work in data collection efforts for this work's sponsor. Conversations were collected using a text-based dialogue interface with the target TOD system. In this way, we collected 1165 human2bot dialogues, 410 interacting with Soloist and 755 interacting with PPTOD. We use these collected dialogues to calculate the human-level GSR, lexical diversity, and syntactic diversity reported in the \textit{Human} column of Table \ref{table:model_comparison}.

\section{Interactive User Simulator}
\label{sec:user_sim}

One of our key goals in the present work is to test the ability of large pretrained language models (LLMs) to generate reasonable, linguistically diverse, and \say{human-like} user-side utterances using only in-context learning in a few-shot setting. The primary motivation for our choice to use in-context learning lies in the fact that, for many novel TOD domains, no annotated human-created dialogue datasets are available for model fine-tuning. Thus, we want to create a user simulator that is able to function without the need for fine-tuning, and that is able to use only a handful of example dialogues to ground its generation. To that end, we take inspiration from the approaches described in recent work on in-context learning with LLMs, such as \citet{kojima2022large} and \citet{wei2022chain} to build a user simulator which is able to effectively engage with extant TOD systems, especially in settings with less complex conversational goals.

We have developed an interactive system which allows our user simulator to converse directly with deployed TOD systems, including end-to-end systems like PPTOD \citep{su2021multi} and Soloist \citep{peng2020soloist}, as well as modular systems such as the ConvLab2 \citep{zhu2020convlab} pipeline TOD system. Additionally, we are able to run our user simulator in a \say{gold} setting wherein the user simulator generates only a single user turn given gold dialogue history data from a target TOD evaluation dataset (such as MultiWOZ \citep{Budzianowski2018MultiWOZA} or SGD \citep{Rastogi2020TowardsSM}), thereby requiring no direct interaction with a deployed TOD system. This latter approach is useful in testing the language generation capabilities of our simulator in isolation. 

\begin{figure}[h]
\includegraphics[width=0.5\textwidth]{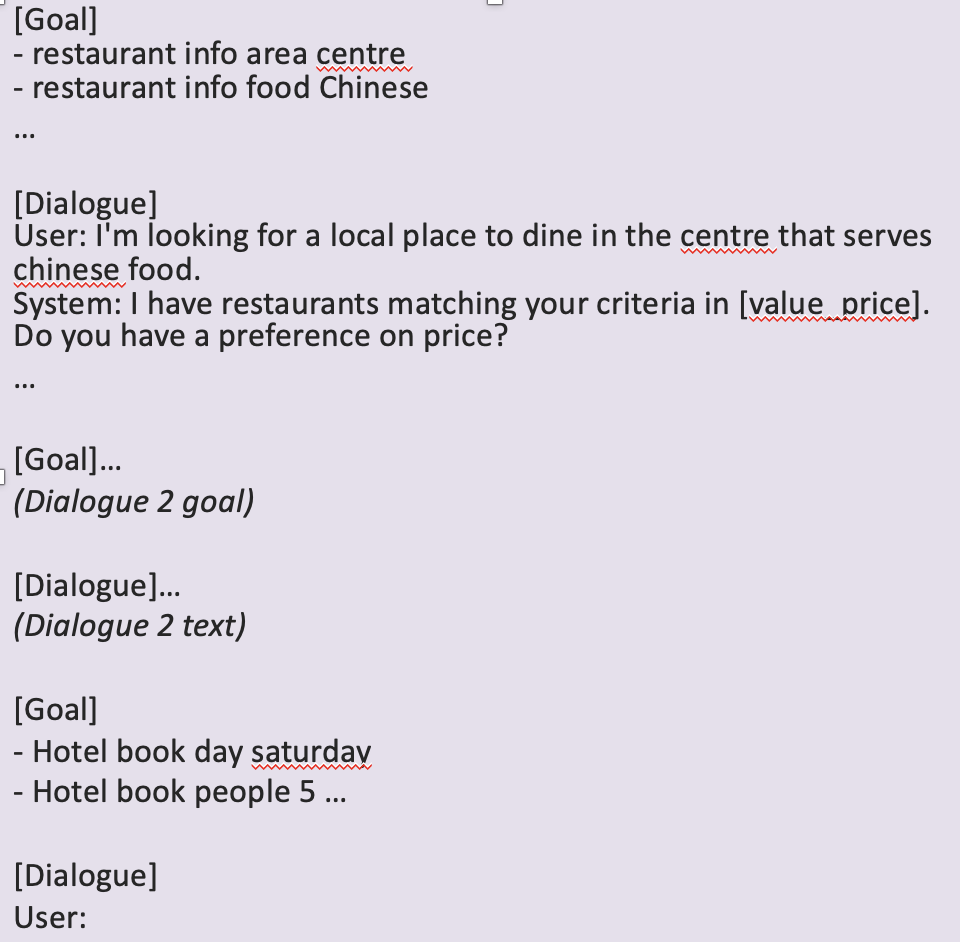}
\caption{Example (abbreviated) initial prompt used to ground user simulator generation at turn 1.}
\label{fig:prompt_design}
\end{figure}

\begin{figure}[h]
\includegraphics[width=0.5\textwidth]{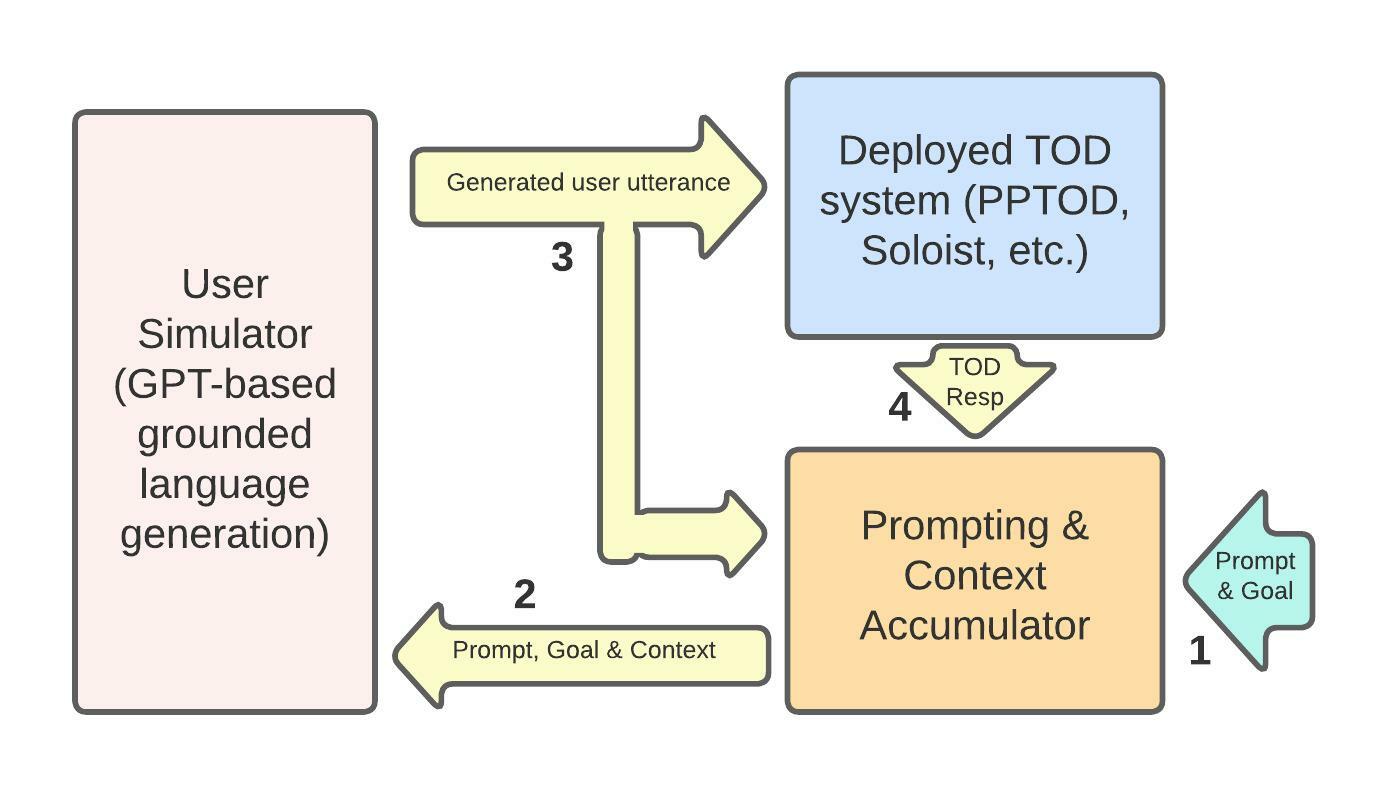}
\caption{Interactive design for testing a TOD system with our proposed user simulator.}
\label{fig:sys_design}
\end{figure}

Our system design, as shown in Figure \ref{fig:sys_design}, consists of a context accumulator and simulator module which formats the provided example dialogues and current conversational goal prompt and queries an LLM to generate a user utterance. This utterance and dialogue history are then provided to the target TOD system to generate a system response. The generated utterance and the system response provided by the TOD system are then added to the context accumulator, and the process repeats until an end of dialogue token is generated by the user simulator. We provide two example conversational goals with their corresponding dialogue histories to the LLM for in-context learning, along with a new goal prompt to initiate the generation of user-side utterances in the target conversation. While we believe that additional examples for in-context learning may improve system performance, we are constrained by the token input limits of the LLM models we are using.  An example initial prompt is shown in Figure \ref{fig:prompt_design}. 

As the user simulator code is based on Huggingface Transformers \citep{wolf2020transformers}, we are able to use a wide variety of generative LLMs as the basis for the simulator. We have primarily tested our system using GPTJ-6B \citep{wang2021gpt} and GPTNeoX-20B \citep{black-etal-2022-gpt}, though we have additionally run limited experiments with GPT-3 davinci \citep{brown2020language} (using the OpenAI API) and AlexaTM-20B \citep{soltan2022alexatm}.

\section{Experiments}

Our driving goal in the present work is to build a user simulator which requires no domain-specific training data and which is able to generate \say{human-like} utterances as measured by goal success, lexical diversity and syntactic diversity. In pursuit of this goal, we tested multiple combinations of prompt design and model selection. In terms of GSR, we found our best performing LLM to be GPT-NeoX-20B, as shown in Table \ref{table:llm_comparison}. As such, we settled on GPT-NeoX-20B to proceed with experiments comparing our user simulator to both human-level performance in TOD interaction and to previous user simulator work. This and all reported experiments were conducted using data from MultiWOZ 2.2 \citep{Budzianowski2018MultiWOZA}

\begin{table}[]
    \centering
    \begin{tabular}{|c|c|}
    \hline
    Model & GSR \\
    \hline
    GPT-J 6B & 11.0 \\
    \hline
    GPTNeoX-20B & 25.0 \\
    \hline
    AlexaTM-20B & 21.2 \\
    \hline
    \end{tabular}
    \caption{Model GSR performance results on MultiWOZ. GPTNeoX performs best and hence is used in the rest of the work.}
    \label{table:llm_comparison}
\end{table}

We similarly tested variation in system performance based on the form of the grounding goal: logical, natural language, and parsed logical form (automatically generated from logical form) using goals drawn from MultiWOZ. Examples of goal grounding settings provided in Appendix B, and results are shown in Table \ref{table:goal_form}. Given these results, we chose to use the parsed logical form in our subsequent experiments. The parsed logical form achieves very similar performance to the natural language form without the need for human-written natural language goals; when creating goals with which to test a TOD system for a new domain, the fact that parsed logical form goals can be automatically generated from the dialogue schema is a distinct advantage. 

\begin{table}
    \footnotesize
    \begin{tabular}{lcccc}
    \hline
    \textbf{Setting} & \textbf{Inform} & \textbf{Success} & \textbf{BLEU} & \textbf{Score} \\
    \hline
    Logical & 54.1 & 24.0 & 7.02 & 46.52\\
    \hline
    NL & 55.0 & 25.0 & 8.07 & 48.07\\
    \hline
    Parsed logical & 55.0 & 25.0 & 7.36 & 47.36\\
    \hline
    \end{tabular}
    \caption {Inform, Success, BLEU and Combo scores achieved by our system on MultiWOZ data in different grounding settings.}
    \label{table:goal_form}
\end{table}
    
Once we established both a model and prompting method for our user simulator, we proceeded to the key aspect of our project experiments: comparing the performance of a user simulator which is implemented using in-context learning in the method described to both human-level performance and the performance of recent works using user simulators based on fine-tuned LLMs to evaluate and test TOD systems. To do this, we first ran all diversity metrics, as described in Section \ref{sec:diversity_metrics} on our Human2bot dataset; these evaluation scripts are being made public along with our user simulator code. We also calculated GSR for the Human2bot data using the MultiWOZ evaluation script provided by \citet{Budzianowski2018MultiWOZA}.

Next we extracted all goals from the MultiWOZ 2.2 test set to use as grounding goals to prompt user-side utterances in our interactive dialogue tests. We then ran our user simulator, as described in \ref{sec:user_sim}, using PPTOD to generate system-side utterances; PPTOD is implemented using the parameters described in \citet{su2021multi}. For our user simulator, we provided two randomly selected dialogues from the MultiWOZ training set as in-context learning data along with the grounding goal, as shown in Figure \ref{fig:prompt_design}. Generation was conducted using GPTNeoX-20B as implemented by Huggingface Transformers. We use the \texttt{generate} method for GPTNeoX-20B with a temperature of 0.5 and a max length of 2048 (the maximum allowed by the model). We tested a range of temperature settings and found 0.5 to offer a reasonable balance between diversity and successful goal grounding. All other parameters for the GPTNeoX model are set to default values. 

We repeated the above process using Soloist as our target TOD system, but with all other experimental parameters the same. We implement Soloist using the parameters described in \citet{peng2020soloist}.

We also setup interactions between the ConvLab2 TUS user simulator and our two target TOD systems (PPTOD and Soloist).  We wrote a simple wrapper to allow the target TOD systems to interact with the ConvLab2 framework, and ran the interaction with the ConvLab2 TUS system using the method described in the ConvLab2 GitHub repository\footnotemark \footnotetext{gitlab.cs.uni-duesseldorf.de/general/dsml/tus\_public}.

The GSR number reported for MetaSim is taken from \citet{Sun2022MetaphoricalUS}, while the lexical and syntactic diversity measures were calculated using our diveristy metric scripts run on MetaSim system output data provided by the MetaSim authors. We were unable to run MetaSim using PPTOD as the target TOD system due to technical issues, thus we do not report PPTOD numbers for the MetaSim user simulator.

\section{Experimental Results}

Our primary experimental results are shown in Table \ref{table:model_comparison}. These results show a comparison between the GSR, lexical and syntactic diversity of human participants (tested using our collected Human2Bot data) and three TOD user simulators (our user simulator, MetaSim, and ConvLab TUS) when interacting with PPTOD and Soloist, respectively, as the target TOD system.

\begin{table*}[hbt!]
    \begin{adjustbox}{width=\textwidth,center}
    \begin{tabular}{lcccccccccccc}
    \multicolumn{1}{c}{} & \multicolumn{3}{c}{\small Human}& \multicolumn{3}{c}{\small Ours}& \multicolumn{3}{c}{\small MetaSim}& \multicolumn{3}{c}{\small ConvLab2 TUS}\\[-.3\normalbaselineskip]
    \multicolumn{1}{c}{} & \multicolumn{3}{c}{\downbracefill} & \multicolumn{3}{c}{\downbracefill} & \multicolumn{3}{c}{\downbracefill} & \multicolumn{3}{c}{\downbracefill}\\
    \hline
    \textbf{Model} & \textbf{GSR} & \textbf{Dep Len} & \textbf{MTLD} & \textbf{GSR} & \textbf{Dep Len} & \textbf{MTLD} & \textbf{GSR} & \textbf{Dep Len} & \textbf{MTLD} & \textbf{GSR} & \textbf{Dep Len} & \textbf{MTLD}\\
    \hline
    PPTOD & 56.9 & 1.98 & 51.25 & 25.0 & 1.8 & 50.3 & - & - & - & 63.7 & 1.4 & 39.2\\
    \hline
    SOLOIST & 71.9 & 1.96 & 50.26 & 23.4 & 1.8 & 48.9 & 80.3 & 1.7 & 49.24 & 60.4 & 1.3 & 37.3\\
    \hline
    \end{tabular}
    \end{adjustbox}
    \caption {Comparison of our proposed method to human, MetaSim, and Convlab2TUS interacting with either PPTOD or SOLOIST. MetaSim and Convlab2 numbers are based on reproduction using our interaction method.}
    \label{table:model_comparison}
    \end{table*}

Table \ref{table:performance_by_intent} shows a detailed breakdown of our user simulator's performance when grounded on goals with different numbers of intents. We include this information to demonstrate the large variation in model performance based on the number of intents in the grounding goal provided to the user simulator. 
    
\begin{table*}[t]
    \begin{adjustbox}{width=\textwidth,center}
    \begin{tabular}{cccccccccc}
    \hline
    \textbf{Num. Intents} & \textbf{Num Dialogs} & \textbf{Num Turns} & \textbf{Inform} & \textbf{Success} & \textbf{BLEU} & \textbf{Combo Score} & \textbf{MTLD} & \textbf{Avg Dep Len} & \textbf{Std Dep Len}\\
    \hline
    1 & 224 & 1057 & 76.79 & 47.77 & 7.79 & 70.07 & 52.46 & 1.86 & 0.77 \\
    2 & 156 & 1148 & 41.67 & 21.79 & 7.67 & 39.4 & 48.75 & 1.8 & 0.69 \\
    3 & 375 & 3000 & 49.87 & 13.33 & 8.54 & 40.14 & 49.92 & 1.81 & 0.72 \\
    4 & 216 & 1878 & 47.22 & 7.87 & 8.09 & 35.63 & 49.92 & 1.88 & 0.76 \\
    5 & 29 & 288 & 41.38 & 3.45 & 8.49 & 30.9 & 47.76 & 1.89 & 0.75 \\
    All & 1000 & 7371 & 53.8 & 20.9 & 8.2 & 45.55 & 50.27 & 1.84 & 0.73 \\
    \end{tabular}
    \end{adjustbox}
    \caption {User simulator performance by intent count on PPTOD. MTLD, Avg Dep Len, and Std Dep Len refer to the MTLD measure of texutal diversity, average and standard deviation of dependency length, respectively}
    \label{table:performance_by_intent}
    \end{table*}
    
Finally, we wished to test the generation abilities of our proposed user simulator at the utterance level. To do this, we test our system in a \say{gold} setting, wherein the system is provided with a dialogue history taken from MultiWOZ and asked to generate the next user turn. Thus, in this setting, there is no interaction with a deployed TOD system. This approach eliminates errors caused by the TOD system (rather than the user simulator) as well as accumulated error in the user simulator. We find that the user simulator is able to achieve a GSR of 77.1\% in this setting, demonstrating that errors from the TOD system side are a major contributing factor to relatively low interactive system performance.
    
\section{Human Evaluation}

While our experimental results show that our proposed user simulator is able to generate lexically and syntactically diverse output, as measured by MTLD and dependency length, we need to further analyze our output to ensure that this diversity is not due to poorly grounded output. In order to further test our proposed user simulator's ability to generate coherent, human-like utterances given a dialogue history, we conduct human evaluation of our simulator output. The goal of this evaluation is to determine how many of the user-side utterances generated by our simulator are considered \say{reasonable} responses by human annotators given the dialogue history. Our annotators indicate that 73.0\% of utterances generated by our system are reasonable given the dialogue history. This high degree of reasonable utterance generation indicates that our increases in diversity and complexity do not come at the cost of coherent generation, further supporting the utility of our proposed system as a reasonable method to test TOD systems, especially those in which the developer does not expect complex, multi-intent goals from users.

Additionally, this qualitative analysis of our model output reveals the primary causes of failed goal completion in the user simulator side of our interactive setup.

\begin{itemize}
    \item Premature termination of dialogues: The user simulator is prone to generating an end dialogue token prior to the successful fulfillment of all intents in the grounding goal. While the user simulator generally starts strong, it frequently terminates the dialogue after the TOD system has fulfilled one or two intents, thereby reducing GSR. This tendency toward partial goal completion is reflected in the \textit{Inform} rates reported in Table \ref{table:performance_by_intent}, which remain relatively stable across intent counts even while GSR rates drop precipitously. 
    
    \item Conversational loops: in the interactive setting, the user simulator and TOD system are also prone to getting stuck in loops with one another, wherein both systems repeatedly respond with the utterance from their respective previous turn. Given that we append both simulator and TOD system generated utterances to the dialogue history to prompt the next round of generation, this problem tends to build upon itself as the in-context learning prompt begins to contain repetitious dialogue, thus encouraging the generation of further repetitious dialogue.
    
    \item LLM hallucination: While less of a problem than the aforementioned, model hallucination is another source of error created by the user simulator. In this situation, the model attempts to achieve intents which are not included in the grounding goal.
\end{itemize}

Example dialogues generated using our interactive user simulator setup, including those exhibiting the degenerate behaviors listed above, are shown in Appendix A. 

\section{Discussion}

One immediate observation when reviewing our primary results presented above is the fact that our in-context learning based user simulator is not achieving a GSR score nearly as high as that achieved by MetaSim and the ConvLab2 TUS simulator. While at first glance this may seem to indicate the the other two systems are better user simulators, this is not necessarily the case. First, as discussed in Section \ref{sec:gsr}, we argue that maximizing absolute GSR is not an appropriate method of evaluating the performance of a TOD user simulator. The goal of a user simulator is not to convey information to the TOD system as effectively as possible, but rather to replicate the behavior of humans interacting with a TOD system. That is, we want the simulator to be \say{human-like}, not just to do a good job talking to a TOD system. As anyone who has worked with TOD systems in actual deployment will tell you, human behavior interacting with these systems is often far from optimal. 

We argue that the high GSR figures reported by the recently developed TOD user simulators, such as MetaSim (reported above) and \citet{Cheng2022IsMA} (86.2 GSR interacting with PPTOD), indicate that these systems are not performing in a truly \say{human-like} manner. 
Our proposed system, while not achieving human-level performance when presented with more complex goals (see Table \ref{table:performance_by_intent}, does does perform well relative to human-level performance on single-intent conversations without any model fine-tuning. Given that a large portion of user goals in real-world interactions with deployed TOD systems consist of single intents, our proposed system is promising as a platform for testing and evaluating newly developed commercial TOD systems. Additionally, our system is able to generate utterances which show near-human level lexical and syntactic diversity and syntactic complexity, furthering its potential utility for testing in-development TOD systems.

As previously mentioned, one of the key benefits of our proposed method is the fact that our approach does not require large amounts of in-domain training data to fine-tune its component language model. This near data-independence makes our proposed user simulator a reasonable choice when testing TOD systems for new domains, for which domain-specific (or at least similar) training data does not exist. To test a system using our user simulator, a developer need only write or collect a few example dialogues and create a set of grounding goals for testing. In the context of a new domain, these goals can either be hand-written or automatically generated from the underlying dialogue schema.

Finally, our proposed user simulator is relatively straightforward and easy to implement, especially compared with systems which consist of multiple interacting modules, such as MetaSim. This makes our proposed method attractive as a way to quickly generate conversations with a new TOD system.

\section{Future work}

While the present work makes significant progress towards effectively simulating user-side dialogue, especially for less complex conversational goals, the use of in-context learning for TOD user simulation remains an open question. As with other in-context learning approaches, the solution space for prompt design is infinite. 
That said, there are several potential improvements to our proposed prompt design which we believe are worth exploring in future work. These include:

\begin{itemize}
    \item Prompting the user simulator to repeat the sub-goal on which it is grounding its output along with the NLG output itself. We believe this may provide additional grounding, thereby improving the system's fidelity to the stated grounding goal.
    
    \item Chain-of-thought prompting:  Recasting the dialogue as a chain-of-thought reasoning problem to take advantage of the properties of GPT-style models discussed in \citet{wei2022chain}.
    
    \item Providing belief state information in the prompts to indicate which goals have been completed during generation. We believe this approach may result in fewer prematurely terminated conversations.
\end{itemize}

There are many other generative LLMs with which we could have tested our approach. For example, we have conducted limited testing with GPT-3 davinci using the OpenAI API, which showed promising results. In future work, we would like to conduct further experiments using larger LLMs, in particular those using reinforcement learning from human feedback, such as ChatGPT, as well as smaller models such as LLaMA.

\section{Conclusion}

In this paper, we have presented a proposed method of using in-context learning with large pretrained language models to create a user simulator for the TOD domain. By conducting extensive tests of the proposed system using data from MultiWOZ, as well as human evaluation, we have shown that our proposed user simulator is capable of generating lexically diverse and syntactically complex output. Additionally, we demonstrate the the proposed system can successfully interact with extant TOD systems, especially when the provided goal is limited to a single intent. While there remains significant room for improvement in the reported results, our system demonstrates that in-context learning alone shows promise in developing effective user simulators for the TOD domain.

\section{Limitations}

One limitation of this paper, and the method proposed herein, is the fact that individuals wishing to use the system to test a new dialogue system must create a goal set for their target domain, which can be a time-consuming process. This fact somewhat limits the \say{few-shot} nature of the proposed method. Additionally, the present paper discusses only a small number of large language models that could be used to implement the proposed method. We understand that there are alternative LLMs that we could have used to test our approach, and we plan to test additional LLMs in future work. That said, we acknowledge that the prompting method we describe may require substantial modification when applied to different LLMs, depending on the target model's training method.

\section{Ethical Considerations}

Whenever using LLMs to generate text, one must consider the fact that such models may be prone to generating biased, erroneous, and harmful content. Given that the method proposed in this paper uses LLMs to test TOD systems in an unsupervised manner, there exists the possibility that the LLM could generate such harmful content in its user simulation. However, users themselves may use harmful and inflammatory language when interacting with TOD systems \say{in the wild}, so the fact that our LLM-based user simulator is capable of generating such language may better reflect real-world user behavior. However, anyone who is interacting with or reviewing data generated by our proposed method should be aware of the potential exposure to harmful content within the generated data.

\bibliography{anthology,custom}
\bibliographystyle{acl_natbib}
\newpage
\onecolumn
\appendix

\section{Appendix A - Example dialogues}
\label{sec:appendix_a}
\begin{itemize}
\item \textbf{Conversational Loop \& LLM Hallucination:}\\
Goal: \{'restaurant': \{'info': \{'food': 'european', 'pricerange': 'expensive'\}, 'reqt': ['address', 'phone', 'area']\}\}
\\\\
User Sim.: I am looking for a place to dine in the expensive price range which serves european food.\\	
TOD Sys.: I have [value\_choice] restaurant -s that meet your criteria . would you like to narrow it down by area ?\\
User Sim.: yes, thank you. a table for 3 at 12:30 on saturday.\\	
TOD Sys.: I have booked you at [value\_name] . your reference number is [value\_reference] .\\
User Sim.: I also want to go to a cinema in the centre.	\\
TOD Sys.: I have [value\_choice] [value\_type] in the [value\_area] . [value\_name] is a great 1 .\\
User Sim.: I will go with that.	\\
TOD Sys.: would you like me to book that for you ?\\
User Sim.: yes, thank you. a table for 3 at 12:30 on saturday.	\\
TOD Sys.: I have booked you at [value\_name] . your reference number is [value\_reference] .\\
User Sim.: I also want to go to a cinema in the centre.	\\
TOD Sys.: I have [value\_choice] [value\_type] in the [value\_area] . would you like me to book 1 for you ?\\
User Sim.: yes, thank you. a table for 3 at 12:30 on saturday.	\\
TOD Sys.: I have booked you at [value\_name] . your reference number is [value\_reference] .\\
User Sim.: I also want to go to a cinema in the centre.	\\
TOD Sys.: I have [value\_choice] [value\_type] in the [value\_area] . would you like me to book 1 for you ?\\
User Sim.: yes, thank you. a table for 3 at 12:30 on saturday.	\\
TOD Sys.: I have booked you at [value\_name] . your reference number is [value\_reference] .\\
User Sim.: I also want to go to a cinema in the centre.	\\
TOD Sys.: I have [value\_choice] [value\_type] in the [value\_area] . would you like me to book 1 for you ?

\item \textbf{Premature Termination:}\\
Goal: \{'hotel': \{'info': \{'name': 'city centre north b and b'\}, 'reqt': ['parking']\}, 'train': \{'info': \{'destination': 'cambridge', 'day': 'sunday', 'arriveBy': '14:00', 'departure': 'broxbourne'\}, 'reqt': ['duration', 'leaveAt', 'price']\}\}
\\\\
User Sim.: I am looking for a hotel in cambridge.	\\
TOD Sys.: there are [value\_choice] [value\_type] in cambridge . do you have a price range or area in mind ?\\
User Sim.: I would prefer to stay in the centre, anything in the moderate price range would be fine.	\\
TOD Sys.: I have [value\_choice] [value\_type] that meet your criteria . would you like me to book 1 for you ?\\
User Sim.: yes please.	\\
TOD Sys.: I would be happy to book you a room . first , can you tell me how many people will be staying , what day you will be arriving , and how many nights you will be staying ?\\
User Sim.: I am travelling with my wife and I will be arriving on sunday and staying for 1 night.\\	
TOD Sys.: I have booked you at the [value\_name] . your reference number is [value\_reference] . is there anything else I can help you with ?\\
User Sim.: no that seems to be everything. thank you!	\\
TOD Sys.: you are welcome . have a great day .\\
User Sim.: bye <end\_dialog>	
TOD Sys.: thank you for using our service today .

\item \textbf{Successful Dialogue:}\\
Goal: \{'train': \{'info': \{'leaveAt': '20:30', 'destination': 'cambridge', 'day': 'friday', 'departure': 'leicester'\}, 'reqt': ['duration']\}\}
\\\\
User Sim.: I would like to book a train from leicester to cambridge on friday.\\	
TOD Sys.: there are [value\_choice] trains that meet your criteria . do you have a time you would like to leave or arrive by ?\\
User Sim.: I would like to leave at 20:30.\\
TOD Sys.: I have train [value\_id] leaving at [value\_leave] and arriving at [value\_arrive] . would you like me to book that for you ?\\
User Sim.: yes, please.	\\
TOD Sys.: how many tickets do you need ?\\
User Sim.: 2	\\
TOD Sys.: I have booked you [value\_people] ticket on [value\_id] . your reference number is [value\_reference] .\\
User Sim.: thank you. <end\_dialog>	\\
TOD Sys.: is there anything else I can help you with today ?

\end{itemize}


\section{Appendix B - Goal grounding settings}
\label{sec:appendix_b}

Examples of our three tested goal grounding settings for the same goal taken from MultiWOZ 2.1

\begin{itemize}
    \item \textbf{Logical Form:}\\
    \{"hotel": \{"info": \{"type": "hotel", "parking": "yes", "pricerange": "cheap", "internet": "yes"\},"book": \{"stay": "3", "day": "tuesday", "people": "6"\}, "fail\_book": \{"stay": "2"\}\}\}

    \item \textbf{Parsed Logical Form:}\\
    - hotel info type hotel.\\
    - hotel info parking yes.\\
    - hotel info pricerange cheap.\\
    - hotel info internet yes.\\
    - hotel book stay 3.\\
    - hotel book day tuesday.\\
    - hotel book people 6.\\
    - hotel fail\_book stay 2.\\

    \item \textbf{Natural Language Form:}\\
    You are looking for a place to stay. The hotel should be in the cheap price range and should be in the type of hotel. The hotel should include free parking and should include free wifi. Once you find the hotel you want to book it for 6 people and 3 nights starting from tuesday. If the booking fails how about 2 nights
\end{itemize}

\end{document}